\documentclass[twocolumn,superscriptaddress,longbibliography]{revtex4-1}
\usepackage{color}
\usepackage[colorlinks=true,urlcolor=blue,citecolor=blue]{hyperref}
\usepackage{soul}
\usepackage{graphicx}
\usepackage{amsmath}
\usepackage{subfigure}
\usepackage{setspace}
\usepackage{hyperref}
\usepackage{url}
\usepackage[hyphenbreaks]{breakurl}
\usepackage{algorithm}
\usepackage{algorithmicx}
\usepackage{algpseudocode}

\sethlcolor{yellow}

\graphicspath{{figures/}}

\makeatletter
\newcommand*\bigcdot{\mathpalette\bigcdot@{.5}}
\newcommand*\bigcdot@[2]{\mathbin{\vcenter{\hbox{\scalebox{#2}{$\m@th#1\bullet$}}}}}
\makeatother

\begin{document}
	
	\title{Quantum-Classical Machine learning by Hybrid Tensor Networks}
    \author{Ding Liu}
    \email{liuding@tiangong.edu.cn}
	\affiliation{School of Computer Science and Technology, Tiangong University, Tianjin 300387, China}

        \author{Jiaqi Yao}
	\affiliation{School of Computer Science and Technology, Tiangong University, Tianjin 300387, China}
    
	\author{Zekun Yao}
	\affiliation{School of Computer Science and Technology, Tiangong University, Tianjin 300387, China}
	
	\author{Quan Zhang}
	\affiliation{School of Computer Science and Technology, Tiangong University, Tianjin 300387, China}

	\begin{abstract}
	Tensor networks (TN) have found a wide use in machine learning, and in particular, TN and deep learning bear striking similarities. In this work, we propose the quantum-classical hybrid tensor networks (HTN) which combine tensor networks with classical neural networks in a uniform deep learning framework to overcome the limitations of regular tensor networks in machine learning. We first analyze the limitations of regular tensor networks in the applications of machine learning involving the representation power and architecture scalability. We conclude that in fact the regular tensor networks are not competent to be the basic building blocks of deep learning. Then, we discuss the performance of HTN which overcome all the deficiency of regular tensor networks for machine learning. In this sense, we are able to train HTN in the deep learning way which is the standard combination of algorithms such as Back Propagation and Stochastic Gradient Descent. We finally provide two applicable cases to show the potential applications of HTN, including quantum states classification and quantum-classical autoencoder. These cases also demonstrate the great potentiality to design various HTN in deep learning way.

	\end{abstract}
	
	\maketitle
	
\section{Introduction.}
In recent years, tensor networks (TN) have drawn more attention as one of the most powerful numerical tools for studying quantum many-body systems \cite{VMC08MPSPEPSRev,O14TNSRev,O14TNadvRev,RTPCSL17TNrev}. Furthermore, TN have been recently applied to many research areas of machine learning \cite{orus2019tensor, efthymiou2019tensornetwork, roberts2019tensornetwork, sun2020tangent}, such as image classification \cite{stoudenmire2016supervised,han2017unsupervised, liu2019machine}, dimensionality reduction \cite{CLOPZ+17TNML1rev, CLOPZ+17TNML2rev}, generative model \cite{han2017unsupervised, Cheng2019Tree}, data compression \cite{li2018shortcut}, improving deep neural network \cite{kossaifi2017tensor}, probabilistic graph model \cite{ran2019bayesian}, quantum compressed sensing \cite{ran2019quantum}, even the promising way to implement quantum circuit \cite{Huggins2018Towards,Benedetti2018A,bhatia2019matrix,ran2019efficient, wang2020quantum}. However, researchers encounter a serious computing complexity problem and raise the question: Are the tensor networks able to be the universal deep learning architecture? As we know, the theoretical foundation of deep neural networks is the principle of universal approximation which states that a feed-forward network with a single hidden layer is a universal approximator if and only if the activation function is not polynomial \cite{leshno1993multilayer, hornik1991approximation}. In this context, the key point of the question of tensor network machine learning is whether TNs can also be universal approximators.

Some pioneering researches have begun to address this fundamental problem. Reference \cite{glasser2018supervised} proposes the concept of generalized tensor networks to outperform regular tensor networks, particularly in terms of representation power. Specifically, they try to combine generalized tensor networks with convolution neural networks together and achieve some good results. In this approach, the convolutions are treated as a feature map, and the tensor network is placed in the final layer and used as the classifier.

Reference \cite{glasser2019expressive} provides a mathematical analysis of the representation power of some typical tensor network factorizations of discrete multivariate probability distributions involving matrix product states (MPS), Born machines, and locally purified states (LPS). Reference \cite{chen2018equivalence} discusses the equivalence between restricted boltzmann machines (RBM) and tensor network states. They prove that these kinds of specific neural networks can be translated into MPS and outline an efficient algorithm for doing so. Drawing on this insight, they quantify the representational power of RBM through the perspective of tensor networks. This insight into tensor networks and RBM guides the design of novel quantum deep learning architectures.

\begin{figure*}[t]
\centering
	% \label{fig:mp:a}
	\includegraphics[width=0.85\textwidth,height=0.25\textwidth]{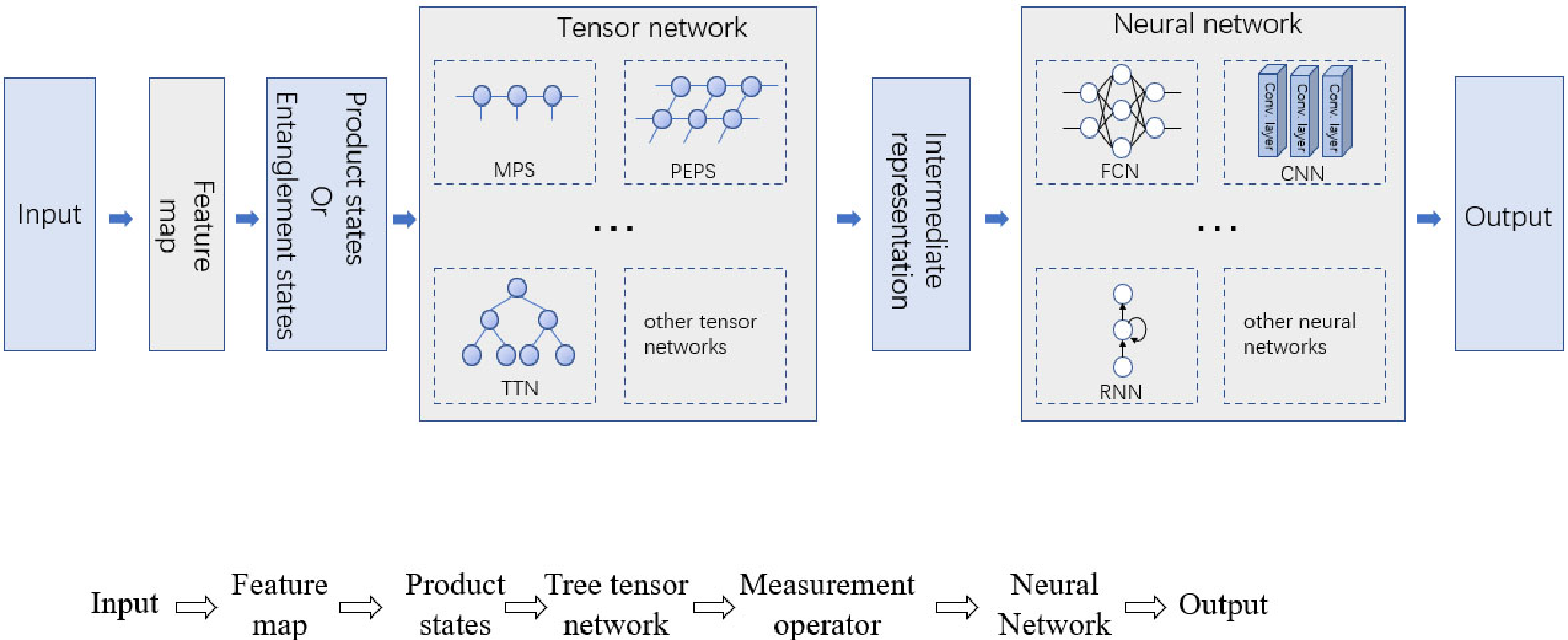}
    \caption{Universal framework of Hybrid Tensor Networks.}
    \label{fig:HTN}
\end{figure*}

\begin{figure*}[t]
	\centering
	\subfigure[]{
		\label{fig:mp:a}
	\includegraphics[width=0.25\textwidth]{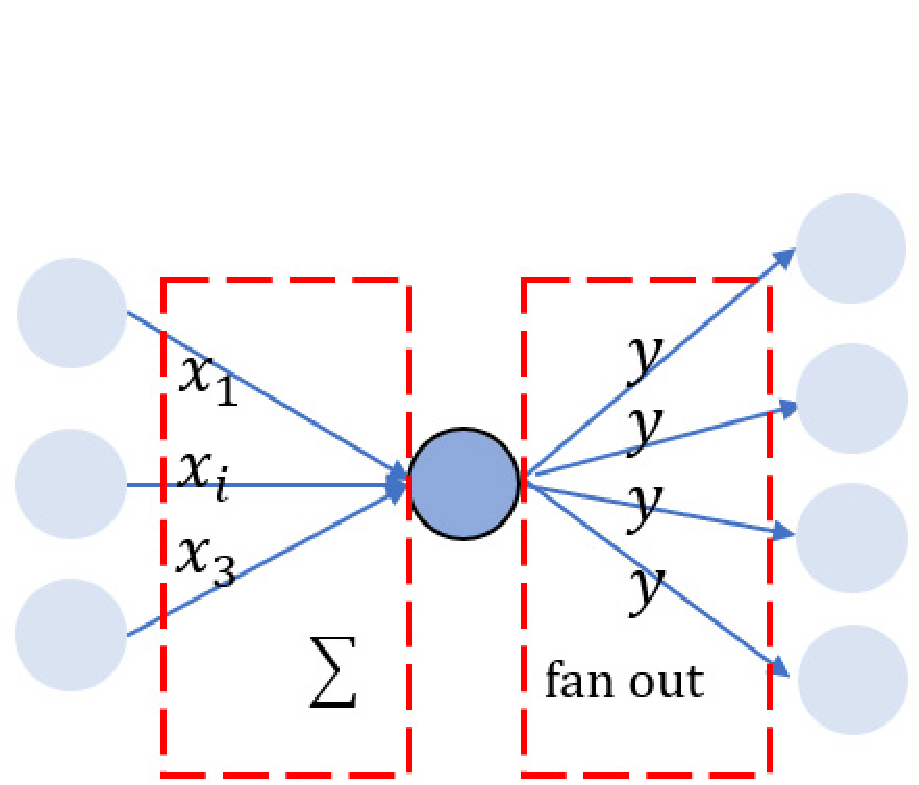}}
	\subfigure[]{
		\label{fig:mp:b}
		\includegraphics[width=0.25\textwidth]{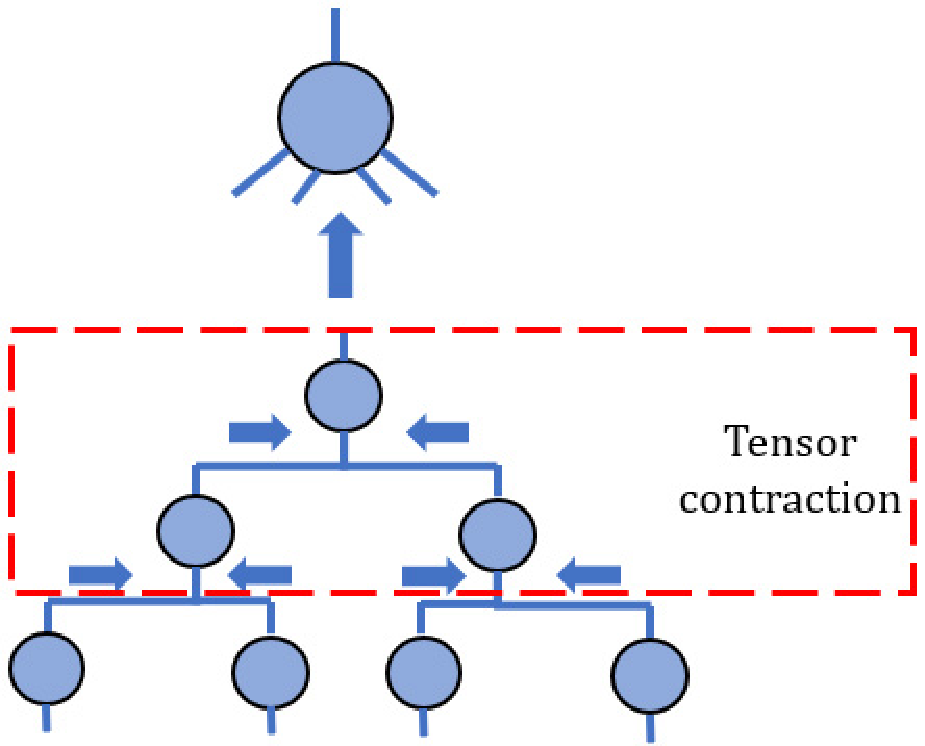}}
	\subfigure[]{
		\label{fig:mp:c}
		\includegraphics[width=0.25\textwidth]{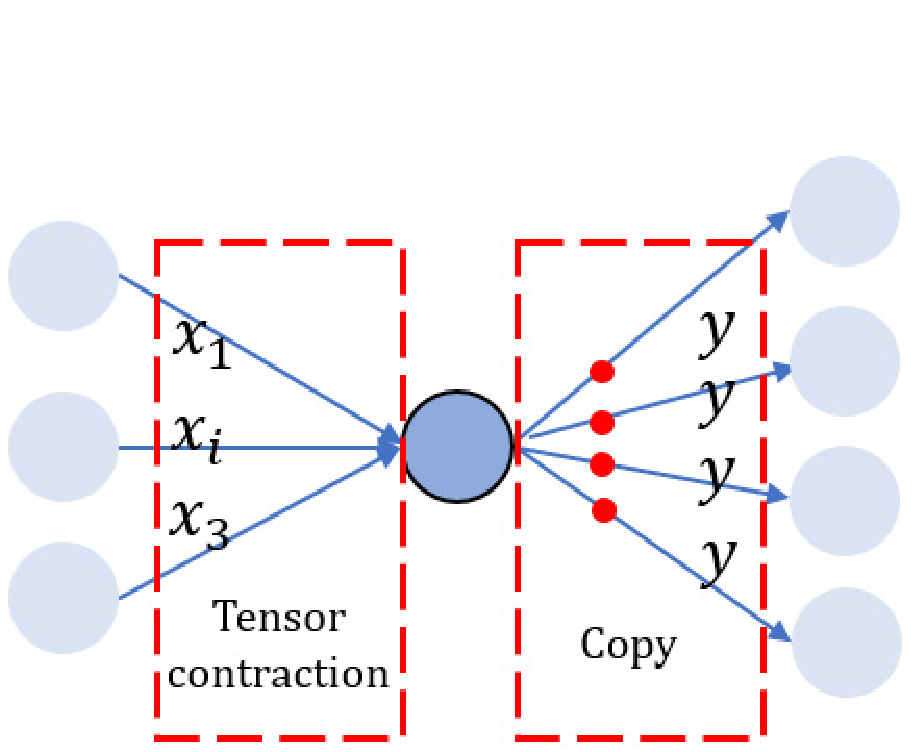}}
	\caption{Different ways of message passing on Neural Networks and Tensor Networks. The directions of message passing are denoted by blue arrows. For tensor network, the message passing is implemented by the operation of tensor contraction; (a) Neural Network ; (b) Regular tensor network; (c) Generalized tensor network, and the operation of copy is marked by red dot; }
	\label{fig:mp}
\end{figure*}

The intersection of physics and machine learning, quantum networks are relatively easy to deploy on quantum computers. Reference \cite{yuan2021quantum} presents the framework of hybrid tensor networks, with building blocks including measurable quantum states and classically contractible tensors. Using hybrid tree tensor networks as an example, it demonstrates the method of efficiently simulating quantum systems on quantum computers with significantly smaller volumes than the target systems. This approach provides insights for simulating large practical problems within medium-scale quantum computers. The review article \cite{rieser2023tensor} discusses how various tensor networks can be mapped to quantum computers, their utilization in machine learning and data encoding, and which implementation techniques can enhance their performance. Additionally, when parameters are randomly initialized, the size of initial gradients decreases exponentially with the increase in the number of quantum bits and circuit depth. To address this phenomenon, known as the "barren plateau", Reference \cite{dborin2022matrix} proposes the MPS pretraining method.

\begin{table}[!htb]
		\centering
			\caption{Number of parameters of each model on MNIST classification.}
			\label{tab:1}       % Give a unique label
			\begin{tabular}{llllllllllll}
				\hline\noalign{\smallskip}
				model                 & MPS  &  TTN   &  LeNet-5 &  FCN   & HTN \\
				\noalign{\smallskip}\hline\noalign{\smallskip}
				Test accuracy         & 98\% &  95\%  &  99\%    &  95\%  & 98\%    \\
				Bond dimension        & 20  &   6    &  -       &  -     & 3       \\
				Number of    &6.3             &  $1.4$           &  $1.2$        &  $2.4$         & $7.7$                      \\
				parameters   &$\times10^{7}$ &  $\times10^9$    &  $\times10^4$ &  $\times10^6$  & $\times10^5$         \\
				\noalign{\smallskip}\hline
			\end{tabular}
\end{table}

\begin{table}[!htb]
	\centering
			\caption{Number of parameters of each model on MNIST regression}
			\label{tab:2}       % Give a unique label
			\begin{tabular}{llllllllllll}
				\hline\noalign{\smallskip}
				Lower bounds of MSE Loss  &  $O(10^{-1})$    &    $O(10^{-2})$    \\
				\noalign{\smallskip}\hline\noalign{\smallskip}
				FCN &  $8.6\times10^3$ & $6\times10^4$ \\
				CNN &  $6\times10^2$    & $3\times10^3$ \\
				TTN &  $4.3\times10^4$  & $1.4\times10^6$ \\
				HTN &  $6.5\times10^3$  & $2.6\times10^4$ \\
				\noalign{\smallskip}\hline
			\end{tabular}
\end{table}

Different from these previous works, we propose the concept of Hybrid Tensor Networks (HTN) which combine tensor networks with classical neural networks into a uniform deep learning framework. We show the schematic of this universal framework in Fig. \ref{fig:HTN}. By virtue of this framework, people are able to freely design HTN by adding any specific tensor network and any classical neural network at any part of the HTN. And then train the whole network using the standard Back Propagation (BP) algorithm and Stochastic Gradient Descent (SGD). Therefore by introducing neurons with nonlinear activation, HTN will be a kind of universal approximator just like neural networks. More importantly, HTN are capable of dealing with both quantum entanglement states and product states. In this way, the HTN will be a good choice for the implementation of a hybrid quantum-classical deep learning model, making it a promising choice for the implementation of hybrid quantum-classical deep learning models. In this paper, we discuss some preliminary ideas to design HTN and provide some applicable cases and numerical experiments. At the end, we give a brief discussion on the quantum feature engineering.

\section{Limitations of regular tensor networks machine learning}\label{sec2}
Although, as a kind of popular and powerful numerical tool in quantum many-body physics, regular tensor networks expose some limitations on machine learning, such as the limitations on representation and scalability in architecture. All of these limitations restrict the application of regular tensor networks in machine learning, especially for deep learning. In this section, we conclude and analyze some main points.

\begin{figure}[H]
	\centering
		% \label{fig:eig:a}
		\includegraphics[width=0.25\textwidth,,height=0.2\textwidth]{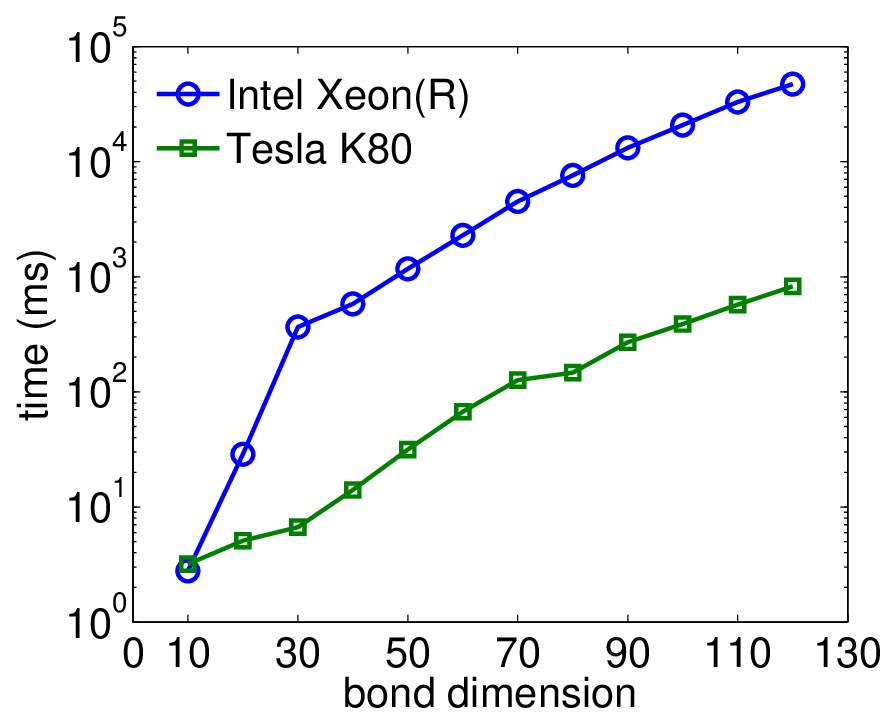}
	\caption{Speeding up on triangle tensor network contraction by GPU platform. The time cost is plotted on logarithmic y-axis.}
	\label{fig:TrTN}
\end{figure}

\begin{figure}[H]
	\centering
	\subfigure[]{
		% \label{fig:mp:a}
		\includegraphics[width=0.22\textwidth,height=0.2\textwidth]{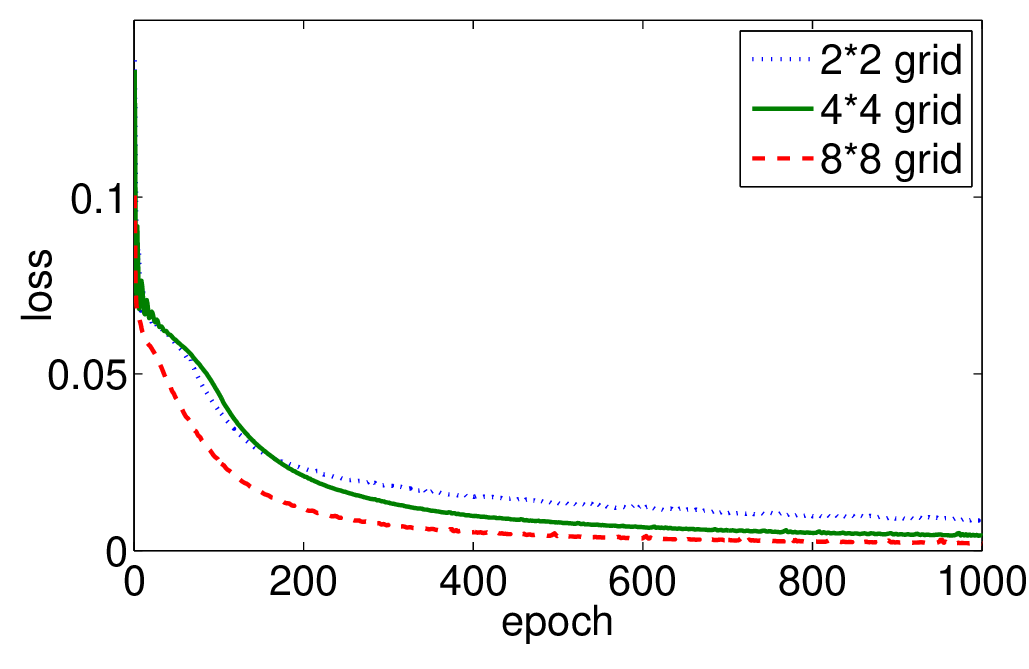}}
	\subfigure[]{
		% \label{fig:mp:b}
		\includegraphics[width=0.22\textwidth,height=0.2\textwidth]{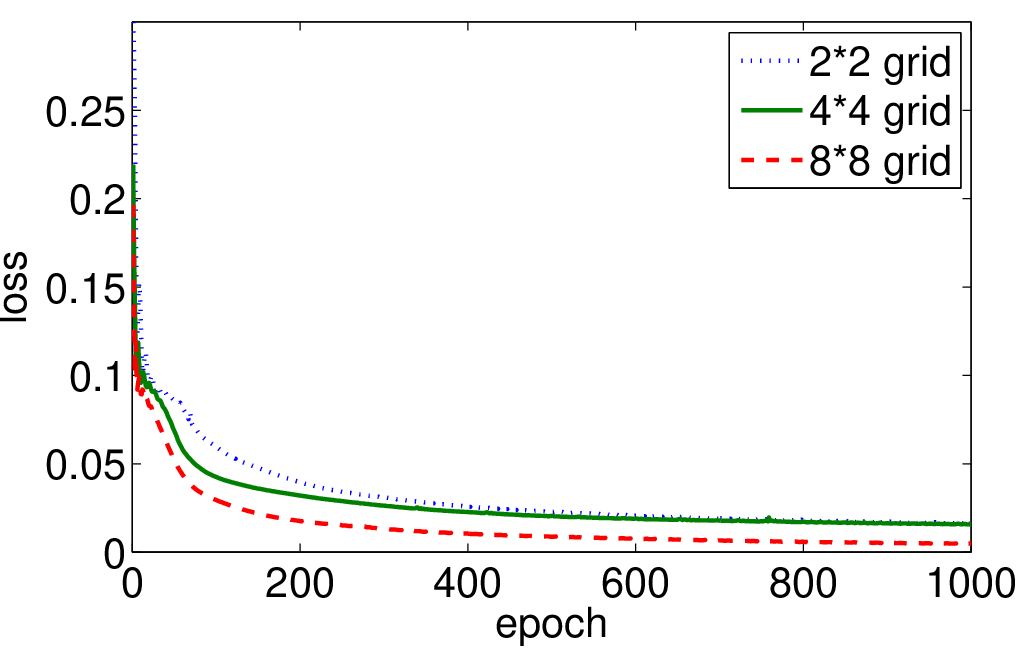}}

	\caption{Training loss of quantum-classical autoencoder; (a) MNIST  ; (b) Fashion-MNIST ;}
	\label{fig:loss}
\end{figure}

\subsection{Representation}
General neural networks (NNs) are characterized by the universal approximation theorem which states that the feed-forward networks are capable of approximating any continuous function, owing to the use of nonlinear activation. So we treat it as a kind of so-called universal approximator. Based on this, NNs become the fundamental building blocks of deep learning. In contrast, TNs are considered as multi-linear functions and therefore obey the superposition principle in quantum mechanics. This is characterized as the intrinsic feature of TNs in quantum many-body systems, but an obstacle to being a powerful universal approximator in machine learning. Therefore, nonlinear feature map functions are required to map all data points from the original feature space to the high-dimensional Hilbert space. In some previous works \cite{stoudenmire2016supervised, liu2019machine}, people use the feature map which is introduced by (\ref{eq-fmap}) firstly as:
	\begin{equation}{C}
	    v_s(x)=\sqrt{\binom{d-1}{s-1}}(\cos(\frac{\pi }{2}x))^{d-{s}}(\sin(\frac{\pi }{2}x))^{s-1}
	\label{eq-fmap}
	\end{equation}
where $d$ denotes the dimension of the physical index, and $s$ runs from $1$ to $d$. By using a larger $d$, the TTN has the potential to approximate a richer class of functions. Furthermore, reference \cite{glasser2018supervised} discusses some other optional complex feature maps, even including neural networks such as CNNs.  These works highlight the crucial role of the feature map in tensor network machine learning, as it determines whether and how well a tensor network can approximate a nonlinear function.

This significant difference between TNs and NNs is illustrated in Fig.  \ref{fig:HTN}. Deep neural networks are equipped with lots of non-linear activation in each layer and each neuron which guarantees its power of approximation. But in contrast, TNs strongly depend on the choice of the feature map, rendering them a multi-linear model in the high dimensional feature space, which strictly limits their power of approximation. It is also easy to understand this from the perspective of statistical machine learning theory such as Support Vector Machine (SVM). In the context of SVM, people always need to map original data points into a high-dimensional feature space and find a kernel function while addressing the nonlinear issue, and it is called the ``kernel trick''. However, in the context of tensor network machine learning, it is unreasonable to endow the TNs with the capacity of universal approximation just by this ``kernel trick'', especially when we want to build a complex and deep tensor network model.

Moreover, for a specific machine learning task, we always have to train a large-scale regular tensor network with more parameters than its corresponding classical neural network can pose a challenge. Taking the previous works as examples \cite{liu2019machine,sun2020generative}, we employed the TTN and MPS on the benchmark of handwritten digits classification. The parametric complexity is around $O(D^5)$ for TTN and $O(D^3)$ for MPS, in which $D$ represents the bond dimension. This may result in the trouble of high parametric complexity since we often need to use a large bond dimension. The experimental results in Table \ref{tab:1} show us the scale of a number of parameters in both TTN and MPS are far more than almost any classical model such as CNN and fully-connected network (FCN). For comparison, we also implement a HTN model and find the number of parameters it needs is less than FCN's, and of course far less than TTN's and MPS's. From the perspective of quantum simulation, we understand that simulating quantum computing on a classical computer always requires exponential growth of parameters with the size of the systems. It shows us a large number of parameters intrinsically leads to a severe problem -–compared with the existing classical deep learning model, it is difficult to train a regular tensor network that has the same or better performance, even if it is impossible.

We also verify this conclusion by some preliminary regression experiments which directly show us how well the model can reach in the curve fitting. Table \ref{tab:2} presents the benchmark results on the MNIST dataset. In this case, we change the classification task to a simple regression issue by setting the label as a corresponding scalar. Taking the class of image ``6'' as an example, we need to train a model that outputs a scalar which closes to ``6'' as soon as possible, rather than a classification vector. We then determine the lower bound of the Mean Square Error loss function (MSE) and find the minimum model that could reach this lower bound. Indeed, the lower bound of the loss function characterizes how well the model can fit the curve. Clearly, the TTN contains many more parameters than FCN and CNN to reach the same level of the lower bound. In this case, it has ranged from around $10$ to $10^3$ times larger than that of the FCN or CNN. It will lead to severe time-consuming problems and even in some worst cases, the training is likely to fail. Similar to the last case, we also find the number of parameters HTN needs is less than FCN's, and far less than TTN's.

It is worth noting that references \cite{novikov2015tensorizing,garipov2016ultimate,ma2019tensorized,li2018shortcut,kossaifi2017tensor} propose tensorizing neural networks or tensor regression networks. These are other feasible ways to take advantage of tensor networks in deep learning. But they are very different from the tensor network learning we talked about here. The motivation for tensorizing neural networks is to compress the weight matrix of neural networks using tensor decomposition to reduce computing complexity or save storage space. In this context, the models of tensorizing neural networks neither take quantum data into consideration nor involve the implementation of a quantum model. 

\subsection{Scalability}
We also evaluate regular tensor networks from the perspective of scalability in architecture design. As we know, there are many deep learning models developed to tackle challenges in computer vision, natural language processing and speech recognition etc., such as the popular CNN \cite{lecun1995convolutional}, RNN, LSTM \cite{hochreiter1997long}, GAN \cite{goodfellow2014generative}, Attention model and Transformer \cite{vaswani2017attention} etc. So just like playing the Jenga game, people are always able to assemble all these models together depending on the engineering applications in practice, even for the designing of extremely deep and complex architecture. In contrast, the scale of tensor networks must be strictly restricted while applying them to quantum many-body systems or machine learning, given the rapid growth of computational complexity. So, it is hard to imagine a huge tensor network with thousands of layers could be simulated on classical computers and applied in machine learning. However, Reference \cite{pan2022solving} proposed that Pan Zhang's team utilized a computing cluster equipped with 512 GPUs, dedicating 15 hours to successfully complete a sampling task of Google's Sycamore quantum supremacy circuit.

\begin{figure}[t]
	\centering
	\subfigure{
	\includegraphics[width=3.4in,height=1in]{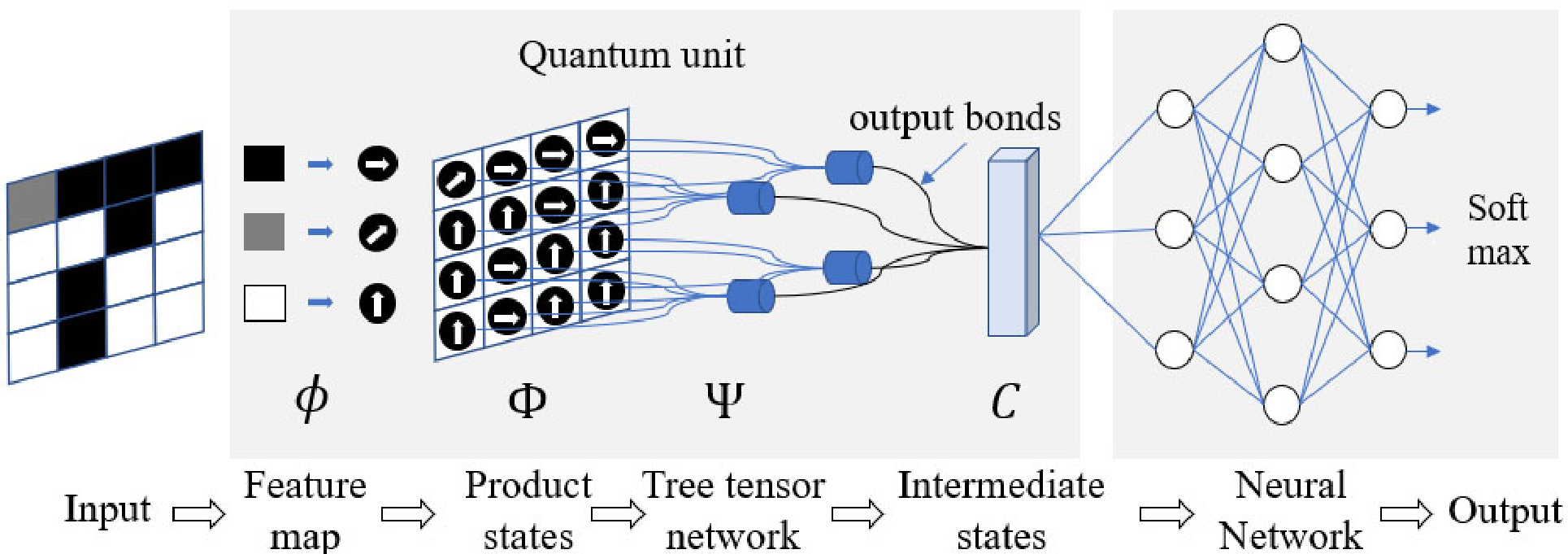}}
	\caption{HTN for quantum states classification. We embed two tree tensor network layers and three dense neural network layers. }
	\label{fig:HTN1}
\end{figure}

\begin{figure}[t]
	\centering
	\subfigure{
	\includegraphics[width=3.4in,height=1in]{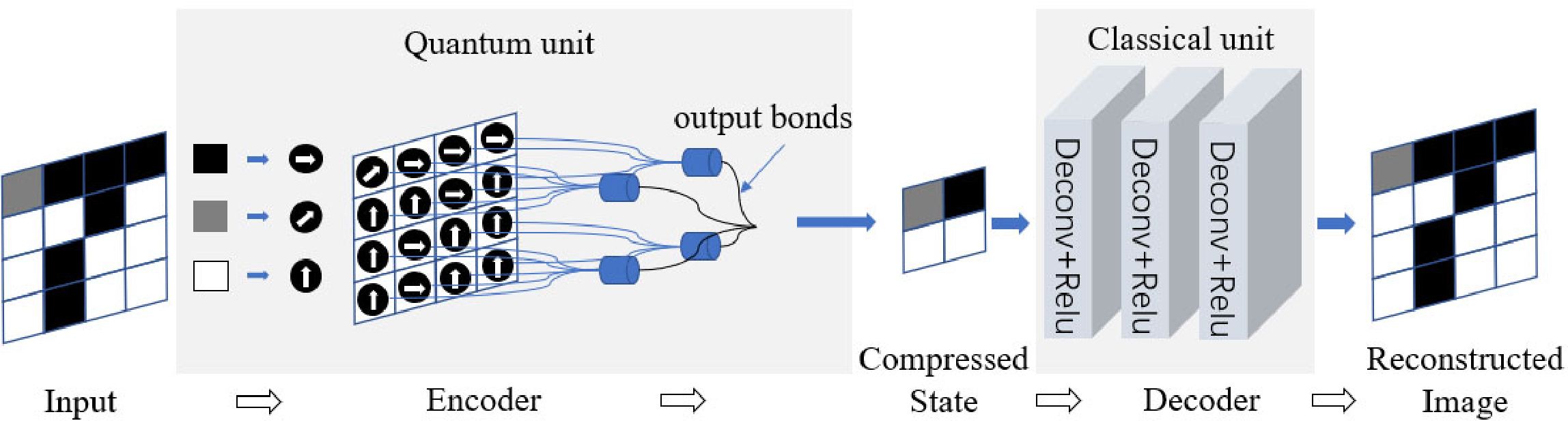}}
	\caption{HTN for Quantum-classical autoencoder. We create the encoder by two tensor network layers, and design three different decoders by using three different setups of deconvolutional layers.}
	\label{fig:HTN2}
\end{figure}

Specifically, we take the process of message passing into consideration and observe a significant difference in behavior between the tensor networks and neural networks. As we show in Fig. \ref{fig:mp:a}, the input message passes through a neural network from the input side to the output side layer by layer. Specifically, for any single neuron, the message passing could be divided into two parts: weighted sum and fan-out. The operation of fan-out generates lots of copies of the output message and distributes them to the next layer. This mechanism guarantees the one-to-many mapping could be implemented easily by neural networks but becomes an obstacle to regular tensor networks for machine learning. Fig. \ref{fig:mp:b} shows that message passing is implemented by contraction in regular tensor networks, and it is essentially the inverse operation of tensor decomposition. Reference \cite{glasser2018supervised} proposed the generalized tensor networks to overcome this limitation by introducing the operation of copy which is marked by the red dot in Fig.\ref{fig:mp:c}.

This mechanism definitely limits the scalability of regular tensor networks in machine learning, especially in the case that we need to build a deep hierarchy.

% \R{In a formal style, we denote the out-degree of a single neuron as $D_{out}$ and the in-degree as $D_{in}$. Then as we know, in a neural network, any neuron is capable of having much larger $D_{out}$ than $D_{in}$, such as fully-connected network . This guarantees the one-to-many mapping could be implemented easily by neural networks. }

% In contrast, Fig. \ref{fig:mp:b} and Fig. \ref{fig:mp:c} show us the process of message passing in two typical tensor networks including MPS and TTN. Since the message passing is implemented by the operation of tensor contraction, thus for a single unit/local tensor, the input message can't be distributed into multi output directions like what neural network can do, unless the operation of tensor decomposition is involved. This mechanism definitely limits the scalability of regular tensor networks in machine learning, especially in the case that we need to build a deep hierarchy.

Based on all these observations, we understand the limitation on architecture scalability is another severe problem for regular tensor network machine learning. Due to this, we think it's not a good way to build a huge, deep and complex deep learning model by regular tensor networks for the practical applications of machine learning. Therefore, how can we take the advantage of tensor networks for deep learning? The solution we try to offer is the Hybrid Tensor Network.

\section{Hybrid Tensor Networks}\label{sec3}
We propose the concept of Hybrid Tensor Networks (HTN) to overcome the limitations of both representation and scalability of regular tensor networks in machine learning. The basic idea is to introduce nonlinearity by the combination of tensor networks and neural networks. By doing so, we are able to embed a local tensor network into any existing popular deep learning framework very easily, involving both model and algorithm such as CNN, RNN, LSTM etc.. Then we could train a HTN by the standard Back Propagation algorithm (BP) and Stochastic Gradient Descend (SGD) which can be easily found in any deep learning literature \cite{goodfellow2016deep}. Suppose we have a HTN which formed in the sequence of $n$ tensor network layers $T_1,T_2,...,T_n$, and subsequent $m$ neural network layers $L_1,L_2,...,L_m$. The cost function is denoted as $Cost$. Then we could  compute the partial derivative to the $ith$ tensor network layer owing to the BP algorithm by  (\ref{eq-bp}).

\begin{equation}{C}
{\frac{{ \partial Cost}}{{ \partial \mathop{{T}}\nolimits_{{i}}}}=\frac{{ \partial Cost}}{{ \partial \mathop{{L}}\nolimits_{{m}}}}\cdot {\frac{{ \partial \mathop{{L}}\nolimits_{{m}}}}{{ \partial \mathop{{L}}\nolimits_{{m-1}}}}\cdot \cdot \cdot {\frac{{ \partial \mathop{{L}}\nolimits_{{1}}}}{{ \partial \mathop{{T}}\nolimits_{{n}}}}\cdot {\frac{{ \partial \mathop{{T}}\nolimits_{{n}}}}{{ \partial \mathop{{T}}\nolimits_{{n-1}}}}\cdot \cdot \cdot \frac{{ \partial \mathop{{T}}\nolimits_{{i+1}}}}{{ \partial \mathop{{T}}\nolimits_{{i}}}}}}}}
\label{eq-bp}
\end{equation}
 Since the operation of tensor contraction defined as (\ref{eq-TNcont}) is doubtless differentiable,

 \begin{equation}{C}
{T\mathop{{}}\nolimits_{{i+1}}^{{ \left[ k \right] }}={\mathop{ \sum }\limits_{{\mathop{{ \alpha }}\nolimits_{{1}}...\mathop{{ \alpha }}\nolimits_{{p}}}}{\mathop{{T}}\nolimits_{{i,\mathop{{ \alpha }}\nolimits_{{1}}}}^{{ \left[ 1 \right] }}\mathop{{T}}\nolimits_{{i,\mathop{{ \alpha }}\nolimits_{{2}}}}^{{ \left[ 2 \right] }}...\mathop{{T}}\nolimits_{{i,\mathop{{ \alpha }}\nolimits_{{p}}}}^{{ \left[ p \right] }}}}}
\label{eq-TNcont}
\end{equation}
where $T_{i+1}^{[k]}$ represents the $kth$ tensor in the $i+1$ layer. So the last term of (\ref{eq-bp}) can be deduced as (\ref{eq-der}),

\begin{equation}{C}
\begin{aligned}
\frac{{ \partial \mathop{{T}}\nolimits^{{ \left[ k \right] }}_{{i+1}}}}{{ \partial \mathop{{T}}\nolimits_{{i}}\mathop{{}}\nolimits^{{ \left[ j \right] }}}}&=\frac{{{\partial \mathop{ \sum }\limits_{{\mathop{{ \alpha }}\nolimits_{{1}}...\mathop{{ \alpha }}\nolimits_{{p}}}}{\mathop{{T}}\nolimits^{{ \left[ 1 \right] }}_{{i,\mathop{{ \alpha }}\nolimits_{{1}}}}...\mathop{{T}}\nolimits^{{ \left[ j \right] }}_{{i,\mathop{{ \alpha }}\nolimits_{{j}}}}...\mathop{{T}}\nolimits^{{ \left[ p \right] }}_{{i,\mathop{{ \alpha }}\nolimits_{{p}}}}}}}}{{ \partial \mathop{{T}}\nolimits_{{i}}\mathop{{}}\nolimits^{{ \left[ j \right] }}}}
\\&={\mathop{ \sum }\limits_{{\mathop{{\{ \alpha }}\nolimits_{{1}}...\mathop{{ \alpha }}\nolimits_{{p}}}\}\backslash \{ \mathop{{ \alpha }}\nolimits_{{j}}\}}
{\mathop{{T}}\nolimits^{{ \left[ 1 \right] }}_{{i,\mathop{{ \alpha }}\nolimits_{{1}}}}...\mathop{{T}}\nolimits^{{ \left[ j-1 \right] }}_{{i,\mathop{{ \alpha }}\nolimits_{{j-1}}}}\mathop{{T}}\nolimits^{{ \left[ j+1 \right] }}_{{i,\mathop{{ \alpha }}\nolimits_{{j+1}}}}...\mathop{{T}}\nolimits^{{ \left[ p \right] }}_{{i,\mathop{{ \alpha }}\nolimits_{{p}}}}}}
\label{eq-der}
\end{aligned}
\end{equation}
where $T_{i}^{[j]}$ represents the $jth$ tensor in the $ith$ layer. And the rest terms of (\ref{eq-bp}) could be calculated easily according to the principle of neural networks. Then, we can update this tensor by using the gradient descend method as (\ref{eq-GD}),

\begin{equation}{C}
{\mathop{{T}}\nolimits_{{i}}^{{'}}=\mathop{{T}}\nolimits_{{i}}- \eta \frac{{ \partial Cost}}{{ \partial \mathop{{T}}\nolimits_{{i}}}}}
\label{eq-GD}
\end{equation}
where $\eta$ denotes the learning rate. Indeed all tensors in HTN could be updated layer by layer following this way. Therefore, it guarantees that the HTN can be trained in the uniform optimization framework which combines BP and SGD. Some popular deep learning open-source software libraries such as Tensorflow \cite{abadi2016tensorflow} and Pytorch \cite{paszke2017automatic} offer powerful automatic differentiation program libraries which could help us implement HTN very easily. 

Furthermore, we test the speedup of tensor contraction on the GPU platform, which is shown in Fig. \ref{fig:TrTN}. It confirms the feasibility of implementing the HTN model by utilizing the GPU platform, and sheds light on the potential, complex and practical applications of large-scale HTN model in the real world.

\begin{figure*}[t]
	\centering
	\subfigure{
		\includegraphics[width=0.85\textwidth]{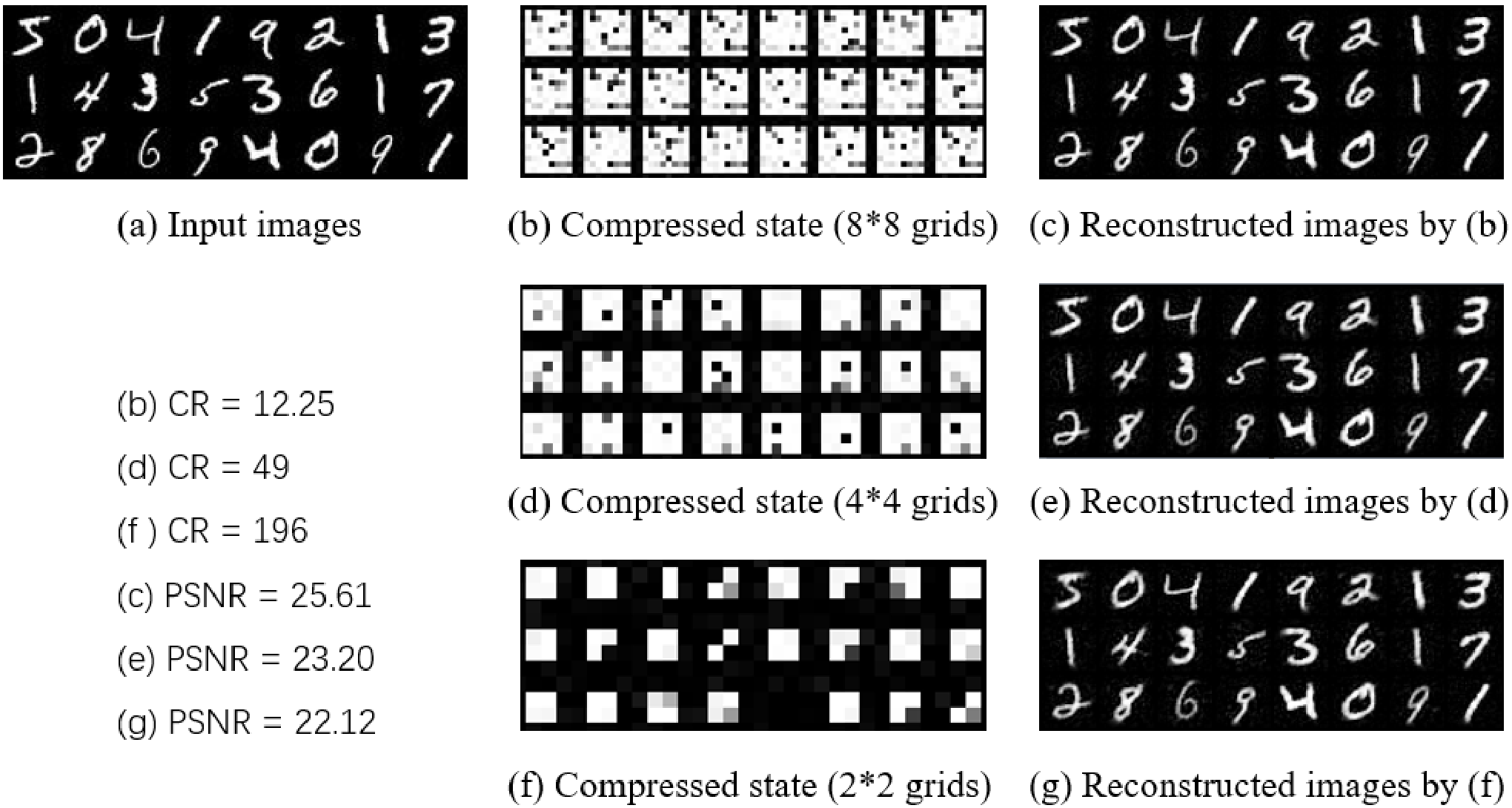}}
	\caption{Quantum-Classical Autoencoder on MNIST;}
	\label{fig:mnist}
\end{figure*}

It is worth noting that, in contrast to previous work that combines tensor networks and neural networks \cite{glasser2018supervised}, we treat tensor networks as the "quantum units" responsible for extracting quantum features from input states. So for the designing of a deep HTN, the first consideration is to determine the role that tensor networks will play. We present our two preliminary attempts on quantum states classification and quantum-classical autoencoder in the following.

\subsection{Quantum states classification}
We design a simple HTN architecture with two tree tensor network layers followed by three dense neural network layers to verify its practicability in classification problems. 
In this case, we first transform the input images into quantum product states without entanglement, which is formed as (\ref{eq-prodstate}), 
\begin{equation}{C}
 \left|\Phi\right\rangle ={ \left| { \phi { \left( {x\mathop{{}}\nolimits_{{1}}} \right) }} \right\rangle } \otimes { \left| { \phi { \left( {x\mathop{{}}\nolimits_{{2}}} \right) }} \right\rangle }\cdot \cdot \cdot \otimes \cdot{ \left| { \phi { \left( {x\mathop{{}}\nolimits_{{n}}} \right) }} \right\rangle }
\label{eq-prodstate}
\end{equation}
where $x_1, x_2, ..., x_n$ represents each pixel; $\left|\Phi\right\rangle$ is the product states we get in the high dimensional Hilbert space; $\phi$ denotes the feature map we mentioned by  (\ref{eq-fmap}). We define the tree tensor network as $\left|\Psi\right\rangle$, so these two tree tensor network layers encode the $\left|\Phi\right\rangle$ into the intermediate low dimensional states $\left|C\right\rangle$ by tensor contraction, i.e. ${{ \left| {C} \right\rangle }={ \left\langle { \Psi } \left| { \Phi } \right\rangle \right. }}$. Afterward, these intermediate states could be processed by neural networks.

\begin{figure*}[t]
	\centering
	\subfigure{
		\includegraphics[width=0.85\textwidth]{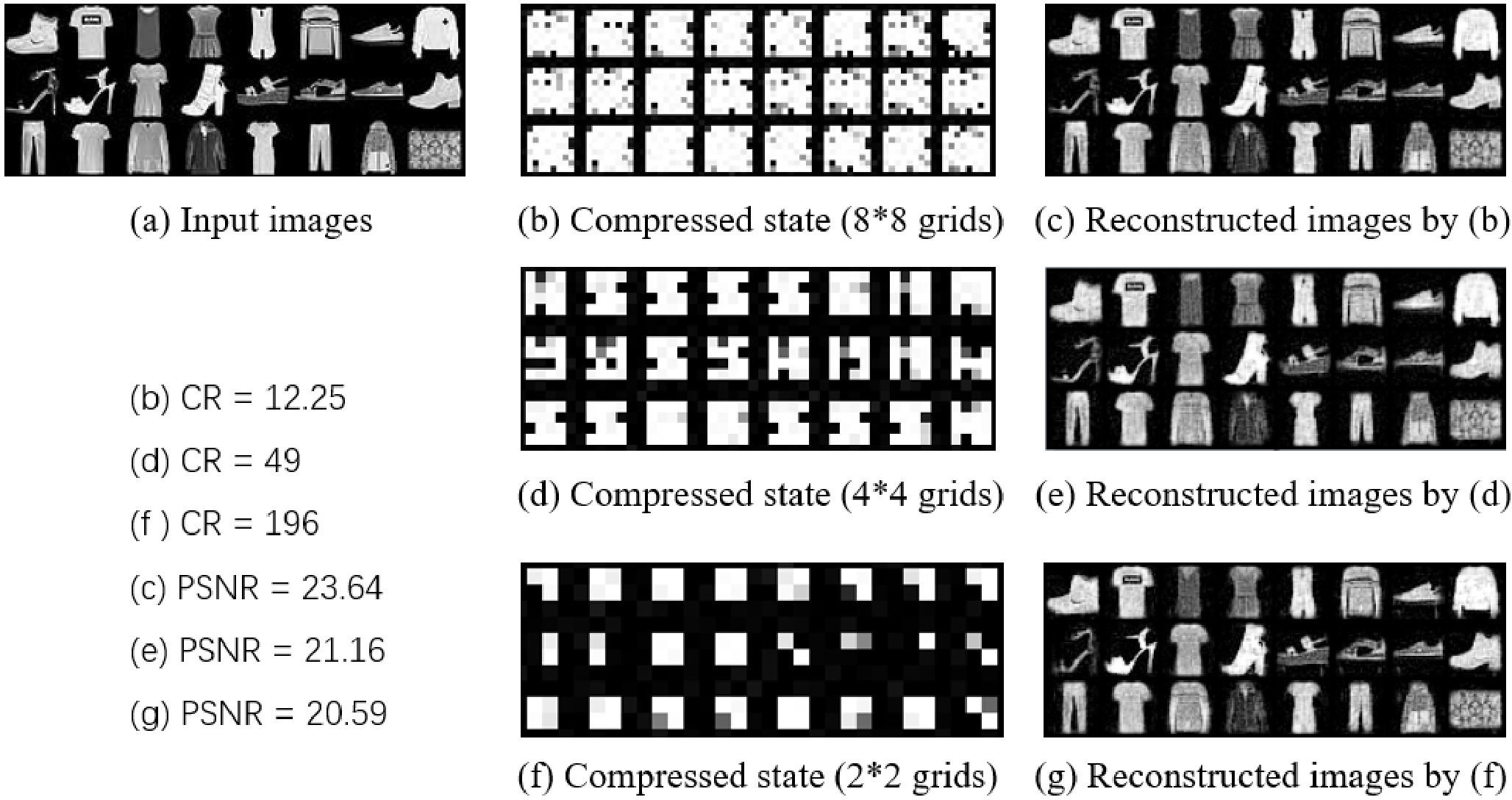}}
	\caption{Quantum-Classical Autoencoder on Fashion-MNIST;}
	\label{fig:fmnist}
\end{figure*}

Finally, the subsequent dense neural network layers classify the intermediate $C$ into 10 corresponding categories by using the cross entropy cost function and the popular Adam training algorithm \cite{kingma2014adam} which is derived from the standard SGD. The cross entropy is defined as (\ref{eq-CroEn})
\begin{equation}{C}
CroEn{ \left( {L,P} \right) }=-{\mathop{ \sum }\limits_{{i=1}}^{{n}}{L \left( C\mathop{{}}\nolimits_{{i}} \left) log \left( P \left( C\mathop{{}}\nolimits_{{i}} \left)  \right) \right. \right. \right. \right. }}
\label{eq-CroEn}
\end{equation}
where $L$ refers to the label and $P$ is the predicted output by HTN. As can be observed, in analogy with the classical CNNs, the tensor network layers play a similar role to the convolutional layers.But different from it, tensor networks are more applicable to quantum state processing because they naturally represent the structure and properties of quantum states. Quantum states typically exhibit highly entangled properties, and tensor networks provide an effective method for describing and processing this entanglement. We benchmark it on the popular MNIST and Fashion-MNIST datasets. The training set consists of 60 000 ($28\times28$) gray-scale images, with 10 000 testing examples. For the simplicity of coding, we rescaled them to (32$\times$32) images by padding zeros pixels. We show the schematic in Fig. \ref{fig:HTN1}. It is easy to get 98\% test accuracy on MNIST and 90\% test accuracy on Fashion-MNIST by using this simple HTN architecture without using any deep learning tricks.

The overview of experimental results for numerous classical models on these tasks can be found on the official websites: of MNIST (http://yann.lecun.com/exdb/mnist/)
and Fashion-MNIST (https://github.com/zalandoresearch/fashion-mnist). Though our method applies to the complex number HTN, we assume all tensors are real for simplicity. Our code of the implementation is available at \cite{htn_code}, and people can find the setup of parameters in detail from it.

\subsection{Quantum-Classical Autoencoder}
We then demonstrate the application of quantum autoencoder by using a variety of HTNs. For simplicity, we still benchmark all models on MNIST and Fashion-MNIST datasets. In this case, the encoder is formed by a tensor network which compresses the input quantum states into low-dimensional  intermediate states. Next, these compressed intermediate states could be recovered by a series of typical classical neural networks. We continue to use the Adam training algorithm, but change the cost function as MSE (Mean Square Error):
\begin{equation}{C}
MSE=\frac{{1}}{{n}}{\mathop{ \sum }\limits_{{i=1}}^{{n}}{ \left( I\mathop{{}}\nolimits_{{i}}-O\mathop{{}}\nolimits_{{i}} \left) \mathop{{}}\nolimits^{{2}}\right. \right. }}
\label{eq-mse}
\end{equation}
where $I$ is the input data, and $O$ denotes the reconstructed data. Fig. \ref{fig:HTN2} shows us the basic architecture and we can find the detailed setup of parameters in our code which is available at \cite{htn_code}.

We show a series of experimental results in both Fig. \ref{fig:mnist} and Fig. \ref{fig:fmnist}, and provide the evaluation indicators Compression Ratio (CR) and PSNR (Peak Signal-to-Noise Ratio) which is defined as (\ref{eq-psnr}):
\begin{equation}{C}
PSNR=10*log\mathop{{}}\nolimits_{{10}} \left( \frac{{max\mathop{{}}\nolimits_{{I}}^{{2}}}}{{\frac{{1}}{{mn}}{\mathop{ \sum }\limits_{{i=1}}^{{m}}{{\mathop{ \sum }\limits_{{j=1}}^{{n}}{ \left( O\mathop{{}}\nolimits_{{i,j}}-P\mathop{{}}\nolimits_{{i,j}} \left) \mathop{{}}\nolimits^{{2}}\right. \right. }}}}}} \right) 
\label{eq-psnr}
\end{equation}
where $max_I$ indicates the max value of input data. 

We compress input product states into intermediate representations in three different scales i.e. 8*8 grids, 4*4 grids and 2*2 grids. It should be noted that larger grids benefit from saving more original input information so that we can reconstruct better images from it and have a better PSNR score. This is evident in both Fig. \ref{fig:mnist} and Fig. \ref{fig:fmnist}. In contrast, the smaller intermediate representations will have higher CR. So it is necessary to strike a balance between them in the practical quantum information application. We also plot the loss curve of both cases in Fig.\ref{fig:loss}, and it clearly shows us the process of training a HTN. Moreover, it definitely shows us the case of 8*8 will reach a lower loss value and produce a better recovery image.

\subsection{Quantum feature engineering}
Above two cases we present the potential of developing the new concept of quantum feature engineering which is the quantum version of feature engineering in machine learning. It is generally recognized that deep learning is an effective way to perform feature engineering since it is capable of extracting feature information automatically from the raw data. Such as during the process of training a convolutional neural network, the convolutional kernels will be trained as feature detectors to recognize, extract and assemble valuable feature information which will be used in the subsequent machine learning tasks such as classification, regression or sequential analysis etc.. In analogy to this standpoint, HTN could be treated as a well-defined hybrid quantum-classical model that is appropriate for quantum feature engineering. 
In which, the tensor network component we say it ``quantum unit'' is in charge of the recognition, extraction or assembling of quantum feature, and then transform it into the form of classical data. 

Although we have no formal definition of what quantum feature exactly is in machine learning, we still started to investigate it involving quantum
entanglement and fidelity by using TTN in our previous work \cite{liu2019machine}. Ref. \cite{youssry2020beyond} proposed the machine learning based approach in terms of quantum control and first proposed the concept of quantum feature engineering. Based on these, we think that quantum feature engineering will be a promising and significant area in quantum machine learning, and we anticipate that HTN will be an excellent choice of quantum feature engineering. In future work, it will help us to understand better how quantum features such as entanglement and fidelity affect the performance of machine learning.

\section{Parameterized Quantum Circuits}\label{sec4}
The article employs Parameterized Quantum Circuits (PQCs) to replicate the Hybrid Tensor Networks (HTN) discussed in the paper, utilizing the PennyLane \cite{bergholm2018pennylane}  and Pytorch packages to construct and simulate the HTN. 
Each pixel is allocated to the corresponding qubit during the PQCs construction, but the MNIST dataset is downsampled to 4*4 grayscale images due to hardware restrictions on the quantum simulation. The specific quantum experiments are segregated into two categories: classification and image reconstruction.

\begin{figure*}[htbp]
	\centering
	\subfigure[]{
		\label{fig:cu:a}
		\includegraphics[width=0.85\textwidth,height=0.75\textwidth]{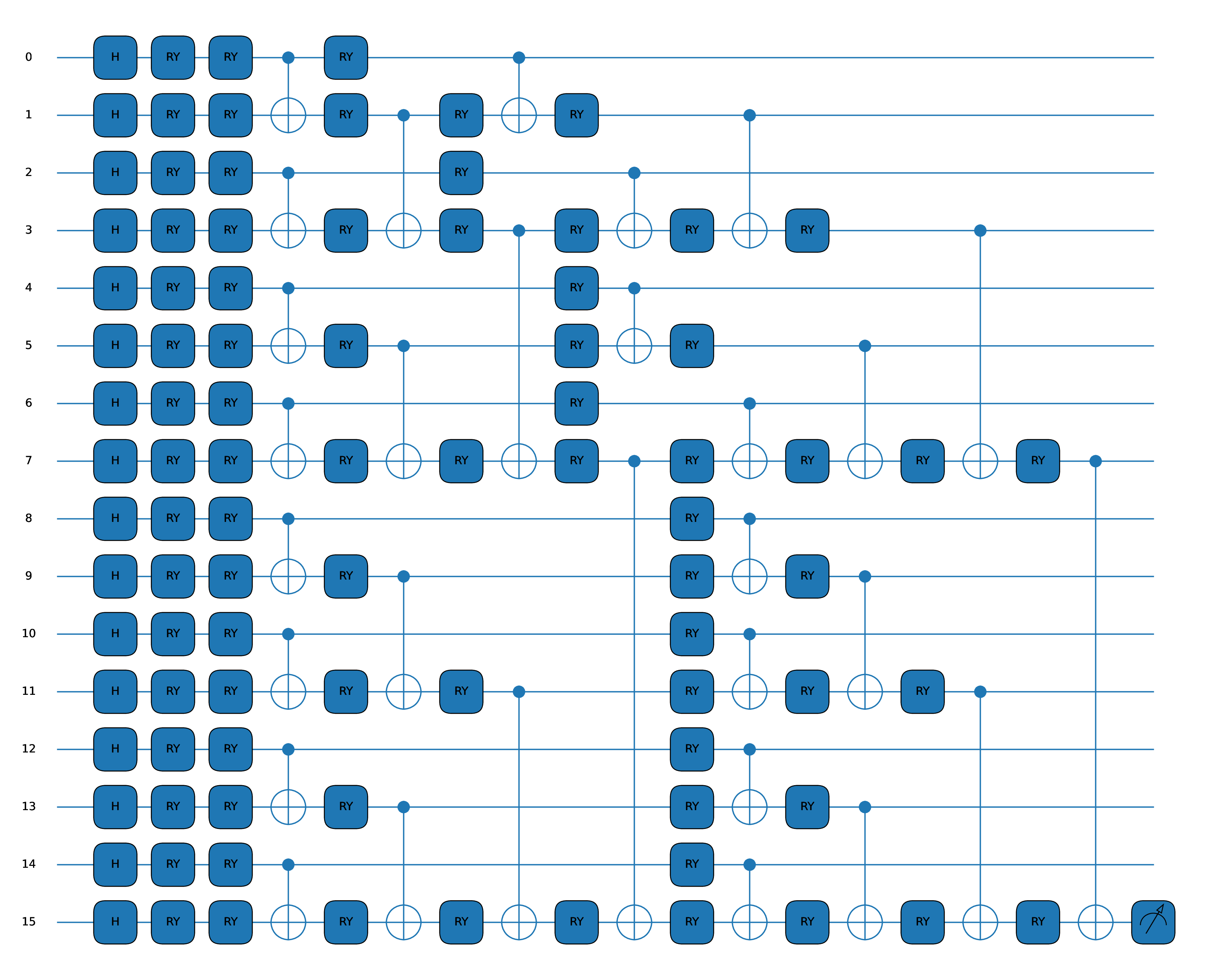}}
	\subfigure[]{
		\label{fig:cu:b}
		\includegraphics[width=0.85\textwidth,height=0.33\textwidth]{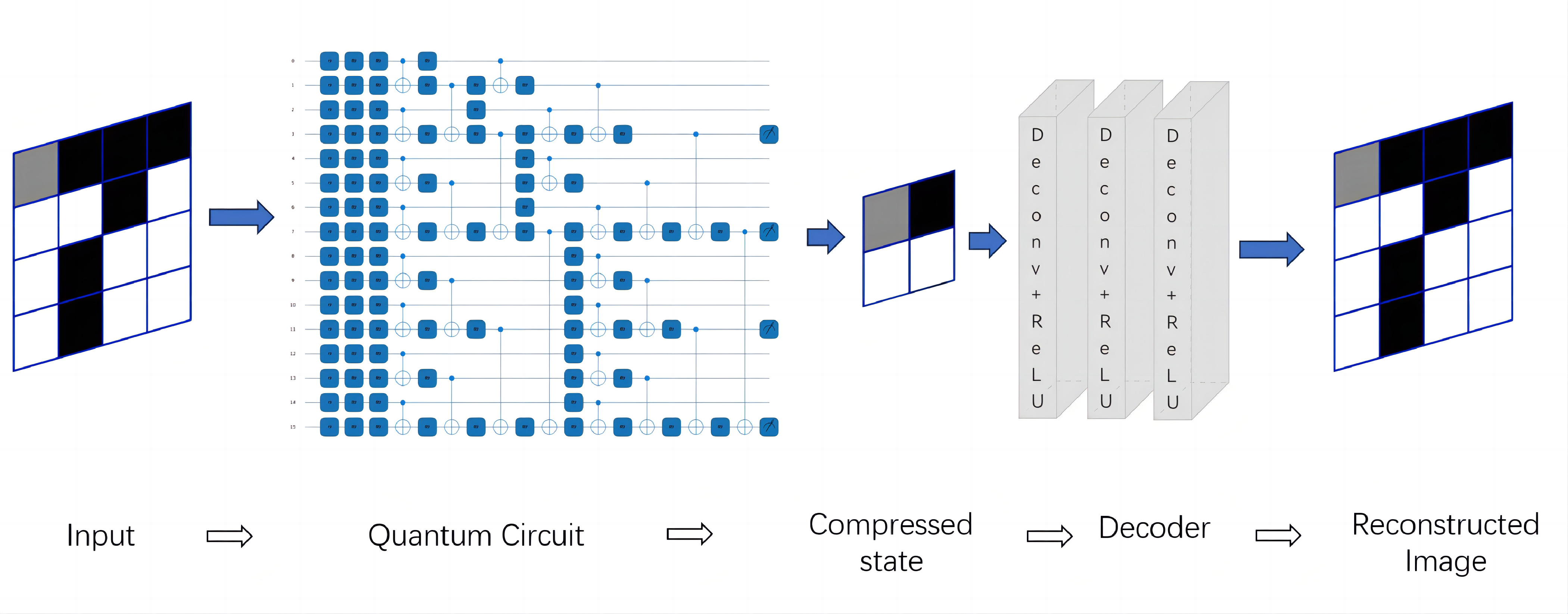}}
	\caption{Quantum circuits of PQCs; (a) Classification  ; (b) Image reconstruction ;}
	\label{fig:Circuit1}
\end{figure*}

\begin{figure}[htbp]
	\centering
	\subfigure[]{
		\label{fig:lo:a}
		\includegraphics[width=0.22\textwidth,height=0.2\textwidth]{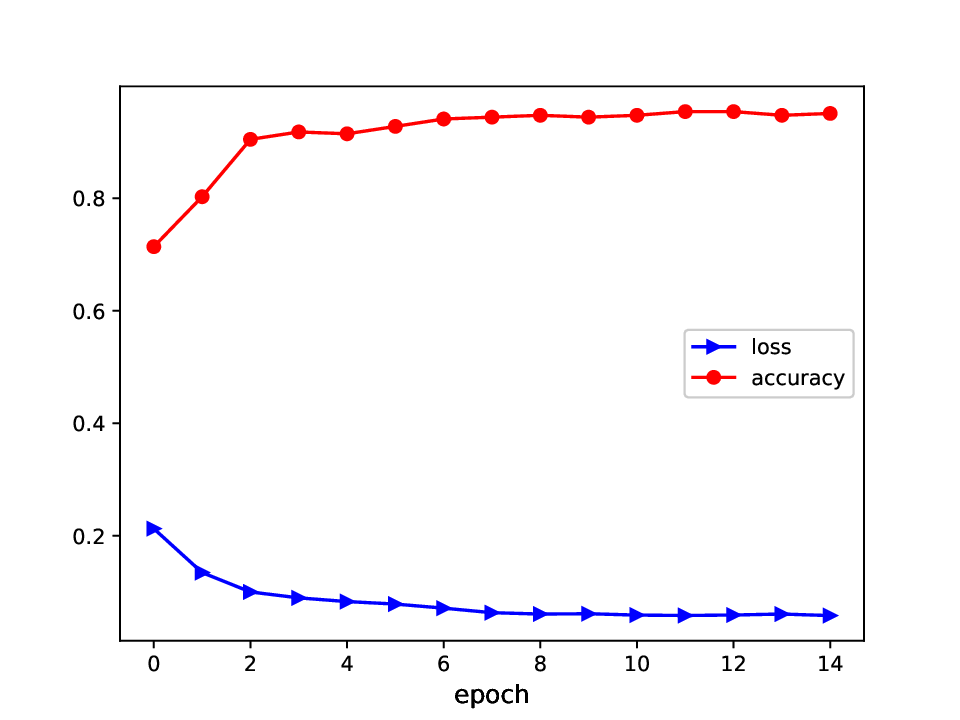}}
	\subfigure[]{
		\label{fig:lo:b}
		\includegraphics[width=0.22\textwidth,height=0.2\textwidth]{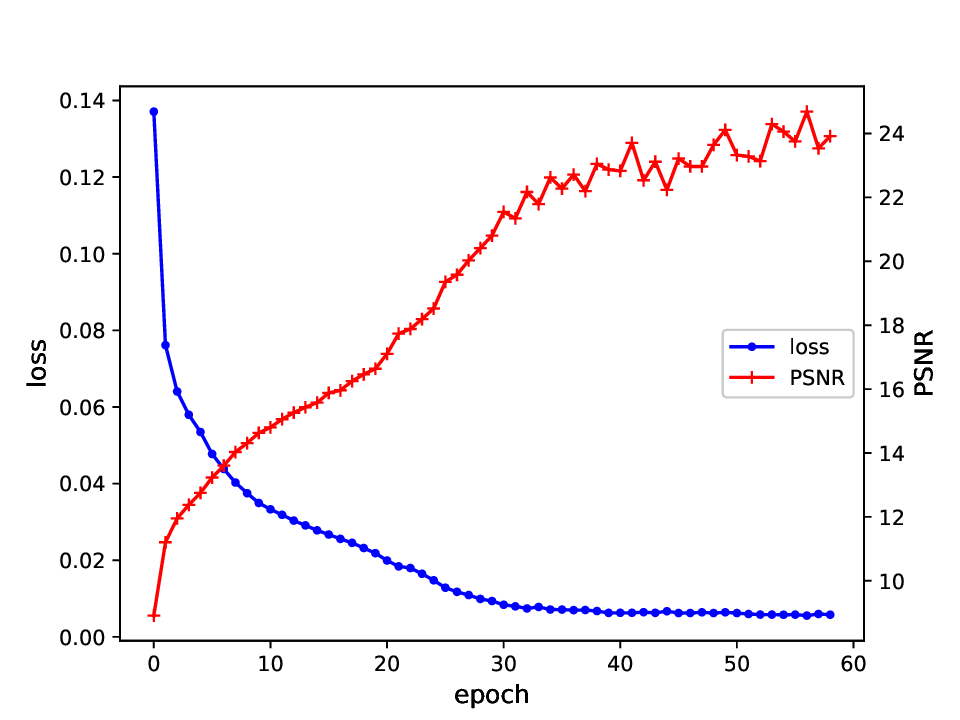}}
	\caption{Training loss of PQCs; (a) Classification  ; (b) Image reconstruction.}
	\label{fig:loss_pqcs}
\end{figure}

\subsection{Classification on sixteen qubits}
The MNIST dataset is downsampled to 16 (4*4 grayscale images). And we select 400 samples from 2 different classes (200/class) to create smaller datasets (from MNIST classes 0, 1). We divide these 400 samples into a train (300) and a test set (100).

Based on the HTN framework presented in Fig. \ref{fig:HTN}, we provide a comprehensive description of the quantum circuit that we implemented. The input stage performs normalization of the 4*4 image, and each pixel corresponds to a qubit. The Feature Map and Product States are constructed using the RY and Hadamard gates to achieve state preparation. The Tree Tensor Network applies the TNN wrapper network in PennyLane for initial model training on the dataset. We use the two-layer TNN function for simulation in this part and choose the RY gate for better fitting ability in network training. We added the measurement function at the end of the quantum circuit that can accept either specified wires or an observable that rotates the computational basis  \cite{bergholm2018pennylane}. Specific PQCs are illustrated in Fig. \ref{fig:cu:a}.

In our experiment, we employed Mean Squared Error (MSE) as the loss function for training PQCs. After 15 epochs, the loss function and accuracy of the network in the training set are shown in Fig. \ref{fig:lo:a}, and we achieved 100\% accuracy on the test set.

\subsection{Image reconstruction on sixteen qubits}
We conduct experiments using the handwritten digit 1 from the MNIST dataset to implement image reconstruction with the Hybrid Tensor Networks (HTN) framework. The specific network model structure can be found in Fig. \ref{fig:cu:b}, and the difference from the classification experiments lies in the measurement stage where we introduce four measurement nodes and reconstruct them into a 2*2 grayscale image. Afterward, we utilize three layers of deconvolution layers to achieve image reconstruction. In this experiment, we use the PSNR  as the evaluation metric for the reconstructed graph. The loss function graph from the experiment is presented in Fig. \ref{fig:lo:b}. We achieved a PSNR of 22.72 on the test set.

\section{Discussion and future work}\label{sec5}
We propose hybrid tensor networks that combine tensor networks with classical neural networks in a uniform framework in order to overcome the limitations of regular tensor networks in machine learning. Based on the numerical experiments, we conclude with the following observations. (1) Regular tensor networks are not competent to be the basic building block of deep learning due to the limitations of representation power i.e. the absence of nonlinearity, and the restriction of scalability. (2) HTN overcomes the deficiency in representation power of the regular tensor network by the nonlinear function from neural network units, and offers good performance in scalability. (3) HTN could be trained by the standard combination of BP and SGD algorithms, allowing for infinite possibilities in designing HTN following deep learning principles. (4) HTN serves as an applicable implementation of quantum feature engineering that could be simulated on classical computers.

There are some interesting and potential research subjects to be left in our future works. The first one is to do deep learning on quantum entanglement data by HTN. Our preliminary experiments in this paper focus on dealing with product quantum states without entanglement, but it is natural to extend HTN to the scenario of quantum entanglement data formed by MPS or PEPS etc., which neural network is incapable of. Moreover, there are some works focusing on tensor network based quantum circuits which demonstrates an interesting way to do quantum machine learning \cite{Huggins2018Towards}. Additionally, some works focus on quantum-classical machine learning by using parameter quantum circuit \cite{xia2019hybrid, otterbach2017unsupervised, zhu2019training,sweke2019stochastic, vinci2019path}. Inspired by these works, the HTN is able to be implemented by parameter quantum circuits in the future. In this case, the training algorithm should be revised to guarantee the isometry of each local tensor in the HTN.

\emph{Acknowledgments}.--- 

DL is grateful to Shi-ju Ran for helpful discussions. And this work was supported by Tianjin Natural Science Foundation of China (20JCYBJC00500) and the Science \& Technology Development Fund of Tianjin Education Commission for Higher Education (2018KJ217).

\bibliography{bibliography}

\end{document}